\documentclass{article}

    \PassOptionsToPackage{numbers, compress}{natbib}

    \usepackage[final]{neurips_2021_preregistration}

\usepackage[utf8]{inputenc} 
\usepackage[T1]{fontenc}    
\usepackage{hyperref}       
\usepackage{url}            
\usepackage{booktabs}       
\usepackage{amsfonts}       
\usepackage{nicefrac}       
\usepackage{microtype}      
\usepackage{xcolor}         
\usepackage{amssymb}
\usepackage{amsmath}
\usepackage{graphicx}
\usepackage{caption}
\usepackage{float}
\usepackage{enumitem}
\title{Enhancing Context Through Contrast}

\author{%
  Kshitij Ambilduke\\
  VNIT, India\\
  \texttt{kshitij.a.vnit@gmail.com}\\
   \And
   Aneesh Shetye\\
   VNIT, India\\
   \texttt{aneeshashetye@gmail.com} \\
   \AND
   Diksha Bagade \\
   VNIT, India \\
   \texttt{diskhabagade003@gmail.com} \\
   \And
   Rishika Bhagwatkar \\
   VNIT, India \\
   \texttt{rishika.vnit@gmail.com} \\
   \And
   Khurshed Fitter \\
   VNIT, India \\
   \texttt{khurshedpf@gmail.com} \\
   \And
   Prasad Vagdargi \\
   Johns Hopkins University, USA \\
   \texttt{prasad@jhu.edu} \\
   \And
   Shital Chiddarwar \\
   VNIT, India \\
   \texttt{shitalsc@mec.vnit.ac.in}
}

\begin{document}

\maketitle
 
\begin{abstract}
Neural machine translation benefits from semantically rich representations. Considerable progress in learning such representations has been achieved by language modelling and mutual information maximization objectives using contrastive learning. The language-dependent nature of language modelling introduces a trade-off between the universality of the learned representations and the model's performance on the language modelling tasks. Although contrastive learning improves performance, its success cannot be attributed to mutual information alone. We propose a novel Context Enhancement step to improve performance on neural machine translation by maximizing mutual information using the Barlow Twins loss. Unlike other approaches, we do not explicitly augment the data but view languages as implicit augmentations, eradicating the risk of disrupting semantic information. Further, our method does not learn embeddings from scratch and can be generalised to any set of pre-trained embeddings. Finally, we evaluate the language-agnosticism of our embeddings through language classification and use them for neural machine translation to compare with state-of-the-art approaches.

\end{abstract}
\section{Introduction}

The performance of Deep Learning models implicitly depends on the data representations \cite{representation_learning_survey}, hence learning paradigms and metrics are defined in ways that optimize the model's capacity to extract useful features from the data. Contrastive Learning (CL) approaches focus on learning representations of data, generally in self-supervised settings \cite{self_sup_contrastive_survey}. The abundance of unlabeled visual data and the ease of introducing subtle yet effective augmentations are two main factors responsible for the success of CL models. The pivotal motivation is to maximize the Mutual Information (MI) between features extracted from augmented views of the data. Although CL paradigms achieve SOTA performance on a variety of tasks, their success cannot be attributed to the properties of MI alone \cite{mutualinfo}.
 
In the lingual domain, representations are affected by the semantic as well as the temporal information present in the data \cite{word_embedding_survey}. Traditional approaches \cite{word2vec,GloVe} try to encode words into vectors according to their relative positions in the corpus, whereas recent approaches optimize performance on language modelling tasks to learn representations \cite{gpt, bert, gpt3, albert, roberta}. Lately, CL-based approaches have emerged for learning universal representations by introducing augmentations during pre-training \cite{contrastive_pretraining_survey, mrasp, mrasp2, infoxlm}. However, the discrete nature of languages makes it difficult to design label-preserving data augmentations \cite{coda}.
Also, training paradigms like Multilingual Masked Language Modelling (MMLM) \cite{bert, infoxlm, xlm} and Translation Language Modelling (TLM) \cite{tlm, infoxlm} require the model to learn language-specific information too \cite{infoxlm}. Due to mixed training objectives, there exists a trade-off between the universality of the learned representations which depends on language-agnosticism and the performance of the model on tasks that require language-specific information \cite{infoxlm}.

Neural Machine Translation (NMT) models aim to translate sentences from one language to another while preserving meaning. This requires models to focus on extracting the semantic and language-agnostic information over the language-specific information. For improving this, we propose a novel Context Enhancement (CE) step that leverages the Barlow Twins loss \cite{barlow} to maximize MI and minimize redundancies between representations of parallel sentences. We do not explicitly augment the data and rather consider sentences as augmented views of their meaning. Further, we do not learn the embeddings from scratch and enhance pre-trained embeddings, increasing generalizability and reducing the compute footprint. Unlike similar works \cite{tlm, infoxlm, xlm}, our objective does not conflict with the primary training objective of NMT. We aim to validate our approach by evaluating performance on Language Classification and NMT by using the WMT-14 \cite{wmt14} En \(\rightarrow\) De and En \(\leftrightarrow\) Fr datasets to compare the performance with state-of-the-art (SOTA) approaches for NMT.

Our main contributions are: \vspace{-3mm}
\begin{enumerate}[leftmargin=*, noitemsep]
    \item Improving performance on NMT through a novel Context Enhancement step that maximizes MI by leveraging a contrastive loss, namely Barlow Twins, without explicitly augmenting the data.
    \item We do not learn embeddings from scratch, hence our method and experiments can be generalised to any set of pre-trained embeddings.

\end{enumerate}

\section{Related works}\label{rel_works}
\subsection{Contrastive learning}
Deep convolutional networks \cite{effnetv2,resnet, inceptionnet, AlexNet, VGG, NFnet, effnet} and even Transformers \cite{vistrans,16x16Trans} have played a foundational role in learning reliable representations from labeled visual data. Owing to the abundance of unlabeled data, there has been a shift from supervised to self-supervised learning \cite{colorizing,jigsaw,context_encoders,context_prediction,image_rotation}. Recently, CL-based approaches \cite{infonce,SimCLR,MoCo,byol,swav,barlow,VICReg} have gained popularity and have shown exceptional performance in a variety of downstream tasks.

Most CL objectives maximize a tractable estimate of the lower bound of MI between two augmented views of the same image \cite{mutualinfo}. Some approaches benefit from large batch sizes \cite{SimCLR} and careful implementation tricks like momentum updates \cite{MoCo, byol} or asymmetric encoders \cite{byol} to prevent collapse. However, Barlow Twins and VICReg \cite{barlow, VICReg} introduce loss functions that naturally avoid collapse and reduce the dependency on the number of negatives while maximizing MI.


\subsection{Neural machine translation}
Performance of Natural Language Processing (NLP) models inherently depends on the word and sentence embeddings. Models trained on large multilingual corpora learn embeddings that can be used for a variety of downstream tasks \cite{bert, sbert, mBART, bart}. Some models try to improve performance on multilingual tasks by focusing on learning language-agnostic components \cite{massively-multilingual, multilingual-nmt, infoxlm, mrasp, mrasp2}. Although centroid subtraction displays signs of eradicating language-specific components \cite{neutralmBERT}, recent works leverage contrastive approaches for the same \cite{infoxlm, mrasp, mrasp2}. However, recent literature has shown that their success cannot be attributed to the properties of MI alone and rather it depends on the choice of feature extractor architectures and the parametrization of the employed MI estimators \cite{mutualinfo}. Due to random masking, MMLM and TLM require the model and embeddings to learn language-specific information \cite{infoxlm}. This may lead to a trade-off between the universality of the learned embeddings and their performance on these tasks. Also, these paradigms require longer training durations \cite{electra} and may not push the model to learn meaningful language semantics by masking common words \cite{xu2020mcbert} or words with too many false negatives \cite{guu2020realm}.

Most approaches for NMT use encoder-decoder architectures \cite{seq2seq, Seq2Seq_with_attention, conv_seq2seq, transformer}. The current SOTA methods \cite{transformer-rev, transformer-pert, noisy-backtranslation, R-drop} introduce subtle yet effective changes in the architecture \cite{transformer-rev} and training method \cite{transformer-pert, R-drop} of the original Transformer \cite{transformer}. Some methods even introduce augmentations by back-translation or by exchanging words with their synonyms and cognates \cite{mrasp, mrasp2, noisy-backtranslation, Transformer-Admin, Data-Diversification}. However, directly augmenting languages may alter the semantic and syntactic correctness \cite{mrasp, mrasp2}.

Since NMT leverages joint information from two sentences, we improve NMT performance by maximizing mutual information and minimizing redundancies between representations of such sentences. Our method does not rely on explicitly augmenting the data and instead treats languages as inherent augmentations introduced in the process of representing abstract meaning. Further, we do not directly maximise any lower bound estimates on the MI but rather use an instantiation of the Information Bottleneck Principle through Barlow Twins \cite{barlow}. In addition, our method does not learn embeddings from scratch but improves the language-agnosticism of pre-trained embeddings.

\section{Approach}\label{sec:approach}
In a typical Transformer-based sequence-to-sequence translation model \cite{transformer}, the encoder \(E(.;\theta_E)\) learns to map a sequence of \(n\)-dimensional word embeddings \(\mathbf{\tilde{x}}=(x_1, x_2, \cdots ,x_{t_1}) \in \mathbb{R}^{t_1\times n}\) from the source language, to a sequence of \(h\)-dimensional latent representations \(\boldsymbol{\tilde{\omega}} =(\omega_1, \omega_2, \cdots ,\omega_{t_1}) \in \mathbb{R}^{t_1\times h}\). This is followed by a decoder \(D(.;\theta_D)\) which maps the sequence \(\boldsymbol{\tilde{\omega}}\), to a sequence of tokens \(\mathbf{\hat{y}} = (\hat{y}_1, \hat{y}_2, ...\hat{y}_{t_2}) \in \mathbb{R}^{t_2\times n}\) in the target language. The encoder uses masked self-attention whereas the decoder uses both cross-attention and self-attention. For a parallel corpus \(\xi\), the loss function \(\mathcal{L}_{trans}\), optimizes the objective \(\mathcal{O}_{trans}\), that is the log probability of obtaining the correct translation \(\mathbf{\tilde{y}}\), given the source sentence \(\mathbf{\tilde{x}}\)

\begin{equation}\label{eqn:1}
    \mathcal{O}_{trans} = -\frac{1}{|\xi|} \sum_{\mathbf{\tilde{x}}, \mathbf{\tilde{y}} \in \xi} \log{p(\mathbf{\hat{y}} | \mathbf{\tilde{x}})}\;\;,\;\;
    \mathcal{L}_{trans} = -\frac{1}{|\xi|} \sum_{\mathbf{\tilde{x}}, \mathbf{\tilde{y}} \in \xi} \mathbf{\tilde{y}}\log{\mathbf{\hat{y}}}
\end{equation}

Unlike recent works \cite{mrasp, mrasp2}, our method does not depend on explicitly augmenting the training data. Rather, we hypothesise corresponding sentences from parallel corpora (\(\mathbf{\tilde{x}}, \mathbf{\tilde{y}}\)) as different views of the same meaning \(\Omega\) i.e. languages are linguistic transforms that map meaning to sentences.

\begin{equation}
    \mathbf{\tilde{x}} = \Lambda_S(\Omega) \;\;,\;\; \mathbf{\tilde{y}} = \Lambda_T(\Omega)
\end{equation}

where, \(\Lambda_S\) and \(\Lambda_T\) represent the linguistic transforms. The encoder tries to learn a transform \(\Lambda_s^\ast\), that maps sentences to their meaning. An ideal encoder-decoder pair would learn the transforms \(\Lambda_s^\ast\) and \(\Lambda_T\) respectively, such that \(\Lambda_s^\ast(\mathbf{\tilde{x}}) = \Lambda_s^\ast(\Lambda_S(\Omega)) = \Omega\) and \(\Lambda_T(\Omega) = \mathbf{\tilde{y}}\).

We intend to improve NMT performance by maximizing MI between the representations of parallel sentences and minimizing the redundant information about the language-specific components. We propose an additional CE step for Transformer-based NMT models that focuses on enriching the language-agnostic features of the sentence embeddings by using a contrastive loss function \(\mathcal{L_{BT}}\) inspired by Barlow Twins \cite{barlow}.

Analogous to the original work \cite{barlow}, we use a Transformer encoder network that encodes two parallel sentences (\(\mathbf{\tilde{x}}, \mathbf{\tilde{y}}\)) from two different languages (\(S, T\)) into two sequences of latent representations \(\boldsymbol{\tilde{\omega}}^S = E(\mathbf{\tilde{x}}; \theta_E)\) and \(\boldsymbol{\tilde{\omega}}^T = E(\mathbf{\tilde{y}}; \theta_E)\). Then a pooling function \(\phi(.)\) is used to obtain sentence embeddings \(\boldsymbol{\sigma}^S = \phi(\boldsymbol{\tilde{\omega}}^S)\) and \( \boldsymbol{\sigma}^T = \phi(\boldsymbol{\tilde{\omega}}^T) \in \mathbb{R}^{B\times h}\). The loss is calculated between batch normalized projections \(\mathbf{Z}^S = \mathtt{BN}(\rho(\boldsymbol{\sigma}^S; \theta_\rho))\) and \( \mathbf{Z}^T = \mathtt{BN}(\rho(\boldsymbol{\sigma}^T; \theta_\rho))\in \mathbb{R}^{B\times d}\) where \(\rho(.;\theta_\rho)\) represents the projection network.

\begin{equation}\label{eqn:2}
    \mathcal{L_{BT}} \triangleq \underbrace{\sum_i \left(1 - \mathcal{C}_{ii}\right)^2}_{\textnormal{\tiny invariance term}} \:+\: \lambda\!\!\!\!\underbrace{\sum_i \sum_{j\neq i} \mathcal{C}_{ij}^2}_{\textnormal{\tiny redundancy reduction term}}
\end{equation}

where \(\lambda\) is a positive constant controlling the relative importance of the two terms. \(\mathcal{C}\) is the empirical cross-correlation matrix computed between the two batches of projections:

\begin{equation}
    \mathcal{C}_{ij}  \triangleq \frac{\sum_b  z_{b,i}^S z_{b,j}^T}{\sqrt{\sum_b(z_{b,i}^S)^2}\sqrt{\sum_b(z_{b,j}^T)^2}} 
\end{equation}

where, \(z_b^S\) and \(z_b^T\) are the \(b\)\textsuperscript{th} batch samples and \(i, j\) indicate the projection network's output dimensions. The cross-correlation matrix \(\mathcal{C} \in \mathbb{R}^{d\times d}\) consists of values between 1 and -1 representing ideal correlation and anti-correlation respectively.

We improve the pre-trained embeddings by optimizing the contrastive loss \(\mathcal{L_{BT}}\) on sentence embeddings with only the encoder during the CE step. For obtaining sentence embeddings, we use pooling as a substitute to the widely used \texttt{[CLS]} token \cite{sbert}. Using the learned weights from the CE step, we attach a matching decoder to train on NMT by optimizing the loss \(\mathcal{L}_{trans}\). However, during the latter step, the encoder's output \(\boldsymbol{\tilde{\omega}}^S\), is directly passed to the decoder without applying pooling, projection or batch normalization.

\begin{figure}[t!]
    \centering
    \includegraphics[width=\textwidth]{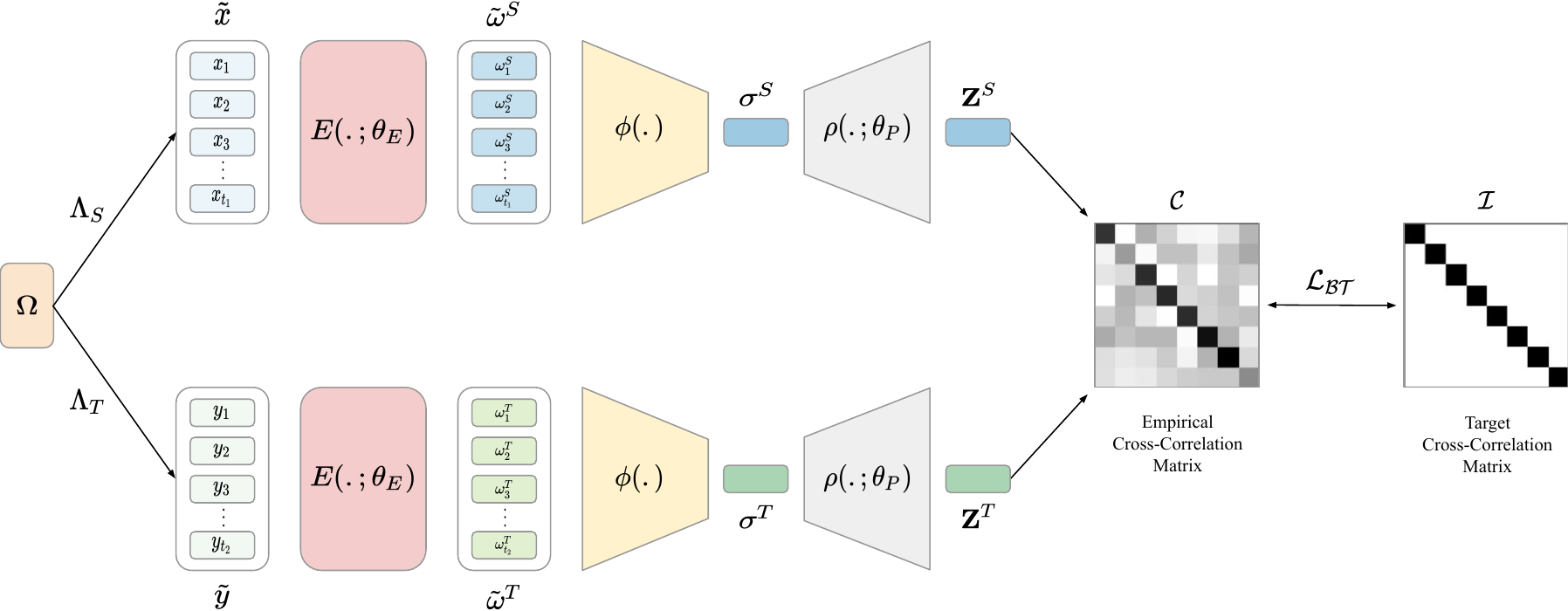}
    \caption{A block diagram of our proposed architecture for the CE step. The encoder maps sentences (\(\boldsymbol{\tilde{x}}, \boldsymbol{\tilde{y}}\)) to sequences of latent representations (\(\boldsymbol{\tilde{\omega}}^S, \boldsymbol{\tilde{\omega}}^T\)). These are then aggregated to get sentence embeddings \(\boldsymbol{\tilde{\sigma}}^S = \phi(\boldsymbol{\tilde{\omega}}^S)\) and \(\boldsymbol{\tilde{\sigma}}^T = \phi(\boldsymbol{\tilde{\omega}}^T)\). The contrastive loss \(\mathcal{L_{BT}}\), is applied to batch normalized projections \(\mathbf{Z}^S = \mathtt{BN}(\rho(\boldsymbol{\sigma}^S; \theta_\rho))\) and \( \mathbf{Z}^T = \mathtt{BN}(\rho(\boldsymbol{\sigma}^T; \theta_\rho))\). The weights \(\theta_E\), are fine tuned for NMT after the CE step. Also, \(\boldsymbol{\tilde{\omega}}^S\) is directly passed to the decoder while training on NMT.}
    \label{fig:1}
\end{figure}

\section{Experiments}
\label{Experiments}

\subsection{Datasets}
\textbf{WMT 2014 English-German:} This dataset \cite{wmt14} contains about 4.5M En-De parallel sentences from Europarl, News Commentary and Common Crawl.

\textbf{WMT 2014 English-French:} This dataset \cite{wmt14} contains about 27.9M En-Fr parallel sentences from Europarl, News Commentary, Common Crawl and the 10\textsuperscript{9} Word corpora.

Further, we will expand our evaluation to other language pairs from distant families following preliminary results.

\subsection{Ablations}
\subsubsection{Context enhancement step} \label{sec:4.2.1}
In the CE step, we use \(N\) encoder blocks of the Transformer to form a sentence encoder. We train the encoder using the Barlow Twins loss \(\mathcal{L_{BT}}\), for a relatively small number of epochs \(\leq 1000\) with different values of \(\lambda\) between 0 and 1 (in steps of 5\(\times\)10\textsuperscript{-3}). We use pre-trained embeddings from mBERT \cite{bert}, InfoXLM \cite{infoxlm}, XLM-RoBERTa \cite{xlmroberta} and XLM \cite{xlm}.

\textbf{Encoder architecture:} We vary the depth of the model \(N\), from 8 to 24 (in steps of 2) and the model dimension \(h\), from about 500 to 2,000 (in steps of \(\approx\) 200), following the work on mBERT \cite{bert}. Further, we vary the number of attention heads from 4 to 12 (in steps of 2).

\textbf{Pooling function:} We experiment with two pooling functions,  \(\phi_{mean}(.)\) and \(\phi_{max}(.)\), representing average pooling and max pooling respectively.

\textbf{Projection network:} The projection network has 3 linear layers, each having \(d\) output units. The first two layers are followed by batch normalization and rectified linear units. We study how the projection dimension \(d\), affects the performance of our model on each evaluation task by varying it from 32 to 16,384 (as per the powers of 2), following the original work \cite{barlow}.


\textbf{Batch size:} We study the dependence of our method on the batch size \(B\), by varying the batch size from 128 to 4,096 (as per the powers of 2), as proposed in the original work \cite{barlow}.

\subsubsection{Translation}
For each of the settings from Section \ref{sec:4.2.1}, we fine-tune the model for NMT after the CE step. A decoder with the same number of layers and model dimension is jointly trained with the context enhanced encoder. Further, the decoder uses masked self-attention and cross-attention as opposed to only self-attention in the encoder \cite{transformer}. However, during translation, the output of the encoder \(\boldsymbol{\tilde{\omega}}^S\), is passed directly to the decoder without using the pooling, projection or batch normalization layers.
\begin{table}[htbp]
\centering
\caption{SACRE-BLEU \cite{sacreBleu} scores (higher is better) on WMT-14 dataset for En-De and En-Fr NMT \\ (\(\dagger\)  Represents methods that use augmentations)}\label{table:1}
\begin{tabular}{lcc}
\toprule
\textbf{Method}     & \textbf{En\(\rightarrow\)De} (\(\uparrow\)) & \textbf{En\(\rightarrow\)Fr (\(\uparrow\))} \\ \midrule
Transformer \cite{transformer}         & 29.12                         & 42.69                \\
MUSE  \cite{MUSE}              & 29.90                         & 43.50                        \\
Depth Growing \cite{Depth_Growing}      & 30.07                         & 43.27               \\
Transformer-Admin \(^\dagger\) \cite{Transformer-Admin}  & 30.10                         & 43.80           \\
Data-Diversification \(^\dagger\) \cite{Data-Diversification} & 30.70                         & 43.70      \\
BERT-Fused NMT  \cite{BERT-Fused}    & 30.75                         & 43.78                  \\
Transformer + RD \cite{R-drop}    & 30.91                         & 43.95                     \\ 
\textbf{Ours (centroid subtracted)}               & -                             & -\\ 
\textbf{Ours (after CE)} &- &- \\
\bottomrule
\end{tabular}
\vspace{-5mm}
\end{table}
\begin{table}[htbp]
\centering
\caption{Tokenized-BLEU scores (higher is better) on WMT-14 dataset for Fr-En NMT.}\label{table:2}
\begin{tabular}{ l  c }
\toprule
\textbf{Method}     & \textbf{Fr\(\rightarrow\)En} (\(\uparrow\))\\ \midrule
Transformer-6 \cite{transformer}       & 39.8                         \\
mRASP2 \(^\dagger\) \cite{mrasp2}              & 39.3                         \\
mRASP \(^\dagger\)  \cite{mrasp}             & 45.4                         \\ 
\textbf{Ours (centroid subtracted)}        & -                            \\
\textbf{Ours (after CE)} & - \\\bottomrule
\end{tabular}
\vspace{-4mm}
\end{table}

\subsection{Evaluation}
For the classification task, sentence embeddings are obtained by pooling the encoder's output. However, for translation, the entire output sequence \(\boldsymbol{\tilde{\omega}}^S\) is passed to the decoder without pooling. 

\textbf{Translation:} We evaluate our model's performance on NMT before and after the CE step. We compare it with SOTA models as shown in Tables \ref{table:1} and \ref{table:2}. Further, we also evaluate the performance after subtracting the centroid from pre-trained word embeddings \cite{neutralmBERT}.

\textbf{Language classification:} To evaluate the language-agnosticism of the embeddings learned by our model, we perform language classification on them. We compute the accuracy \(a_1\) of a language classifier \(C_1\) trained on sentence embeddings obtained from mBERT after pooling. Freezing the parameters of \(C_1\), we evaluate it's accuracy \(a_2\) on embeddings obtained after the CE step. Then, we train a language classifier \(C_2\) on embeddings obtained after the CE step and compute it's accuracy \(a_3\). For both word and sentence embeddings, the relation \(a_2 < a_3 < a_1\) indicates the absence of language-specific components in the embeddings, validating an increase in language-agnosticism.


To compare our method with prior works \cite{neutralmBERT}, we compute \(a_1^\prime, a_2^\prime\) and \(a_3^\prime\) before and after subtracting the centroid of all sentence embeddings.
We compute \(a_1^\prime\) by training a classifier \(C_1^\prime\) on the sentence embeddings obtained from mBERT. Then, for computing \(a_2^\prime\), we evaluate \(C_1^\prime\) on the centroid subtracted sentence embeddings. Finally, we compute \(a_3^\prime\) as the accuracy of a model \(C_2^\prime\) trained on the centroid subtracted sentence embeddings. We extend this entire procedure for word embeddings too.

\textbf{Qualitative analysis:} 
We visualize the distribution of word and sentence embeddings using t-SNE plots of word and sentence embeddings before and after the CE step. To analyse the word-level redundancies, we plot the correlation matrices between corresponding word pairs at different stages of the CE step. Further, we plot the attention maps of every head of the encoder and decoder to evaluate how the CE step affects the attention mechanism.

\section{Results}
We followed the experimental protocol proposed in the above sections Sec. \ref{Experiments}. We tested our code and hypothesis on smaller datasets like opus\_rf \cite{} before deploying our code on WMT-14. However, we were unable to produce feasible results due to the following reasons

\subsection{Incompatibility of datasets}
The current version of the WMT-14 datasets is incompatible with the current versions of PyTorch.

\subsection{Mode collapse due to BERT, XLM-RoBERTa}
As proposed in the section \ref{sec:4.2.1}, to test our hypothesis we first trained a transformer model to perform translation. This was followed by the context enhancement step. After completion of CE step, we fine tuned the transformer model from the first step to again perform translation. We used the embeddings from BERT and XLM-RoBERTa which resulted in collapsed embeddings after the completion of initial translation. Performing CE step on such collapsed embeddings did not result in improved performance.

\subsection{No improvement even while training from scratch}


\section{Conclusion}

We propose a novel Context Enhancement step to push neural machine translation performance using contrastive learning. Our method maximizes Mutual Information between two views of the same meaning by leveraging the Barlow Twins loss. Unlike most works, our method does not depend on explicit augmentations or implementation tricks. Further, our proposed objective pushes the model to learn language-agnostic features which directly improves neural machine translation performance.




\bibliography{ref.bib}
\end{document}